\documentclass[10pt,twocolumn,letterpaper]{article}

\usepackage{cvpr} 

\usepackage{times}
\usepackage{epsfig}
\usepackage{graphicx}
\usepackage{amsmath}
\usepackage{amssymb}
\usepackage{booktabs}
\usepackage{multirow}
\usepackage{array}
\usepackage{xcolor}
\usepackage{microtype}
\usepackage{float}
\usepackage{subcaption}
\usepackage{enumitem}
\usepackage{url}
\usepackage{hyperref}
\begin{document}

\title{DualSwinFusionSeg: Multimodal Martian Landslide Segmentation via\\
Dual Swin Transformer with Multi-Scale Fusion
and UNet++}

\author{
Shahriar Kabir$^{1,*}$, Abdullah Muhammed Amimul Ehsan$^{1,*}$,
Istiak Ahmmed Rifti$^{1,*}$ \\ Md Kaykobad Reza$^{2}$ \\
$^{1}$Bangladesh University of Engineering and Technology, Bangladesh\\
$^{2}$University of California, Riverside, USA\\
{\tt\small kabirshahriar468@gmail.com \quad amimul.ehsan2001@gmail.com}\\
{\tt\small riftirc96@gmail.com \quad mreza025@ucr.edu}\\
{\small $^{*}$Equal contribution}
}

\maketitle
\thispagestyle{empty}

\begin{abstract}
Automated segmentation of Martian landslides, particularly in tectonically active regions such as Valles Marineris,is important for planetary geology, hazard assessment, and future robotic exploration. However, detecting landslides from planetary imagery is challenging due to the heterogeneous nature of available sensing modalities and the limited number of labeled samples. Each observation combines RGB imagery with geophysical measurements such as digital elevation models, slope maps, thermal inertia, and contextual grayscale imagery, which differ significantly in resolution and statistical properties. To address these challenges, we propose \textbf{DualSwinFusionSeg}, a multimodal segmentation architecture that separates modality-specific feature extraction and performs multi-scale cross-modal fusion. The model employs two parallel Swin Transformer V2 encoders to independently process RGB and auxiliary geophysical inputs, producing hierarchical feature representations. Corresponding features from the two streams are fused at multiple scales and decoded using a UNet++ decoder with dense nested skip connections to preserve fine boundary details. Extensive ablation studies evaluate modality contributions, loss functions, decoder architectures, and fusion strategies. Experiments on the MMLSv2 dataset from the PBVS 2026 Mars-LS Challenge show that modality-specific encoders and simple concatenation-based fusion improve segmentation accuracy under limited training data. The final model achieves 0.867 mIoU and 0.905 F1 on the development benchmark and 0.783 mIoU on the held-out test set, demonstrating strong performance for multimodal planetary surface segmentation.
\end{abstract}

\section{Introduction}
\label{sec:intro}

Landslides are among the most significant mass-wasting processes shaping the surfaces of rocky planetary bodies. On Mars, they modify crater walls, polar scarps, and volcanic flanks, providing valuable geomorphological evidence of subsurface volatiles and potential past liquid activity. Accurate mapping of these features is important for planetary geology, hazard assessment, and future robotic exploration. However, manual annotation of landslides from orbital imagery is time-consuming and difficult to scale to planetary-level analysis. Automated segmentation methods offer a practical solution but remain challenging due to severe class imbalance, large morphological variability across terrains, and the heterogeneous nature of multi-sensor observations.

The PBVS 2026 Martian Landslide Segmentation (MARS-LS) Challenge addresses this problem using a multimodal remote sensing dataset designed for binary landslide segmentation. Each $128\times128$ GeoTIFF tile contains seven co-registered channels, including three  RGB bands and four geophysical modalities: Digital Elevation Model (DEM), slope, THEMIS thermal inertia, and CTX grayscale context imagery. These modalities span dramatically different spatial resolutions, ranging from approximately 6 m for CTX grayscale imagery to approximately 232 m for the Viking RGB mosaic. Furthermore, only 531 labeled tiles are available for training and validation, placing the problem firmly in a small-data regime where effective multimodal feature representation and robust architectural design are critical.

To address these challenges, we propose \textbf{DualSwinFusionSeg}, a multimodal segmentation framework that explicitly separates modality-specific feature extraction while enabling efficient cross-modal interaction. The model employs two parallel Swin Transformer V2-Small encoders that independently process RGB imagery and auxiliary geophysical inputs, producing hierarchical feature representations. Features from the two streams are fused at multiple pyramid levels using lightweight concatenation-projection modules and decoded using a UNet++ architecture with dense nested skip connections to preserve fine spatial details. To better understand design trade-offs under limited data, we conduct extensive ablation studies covering modality combinations, loss functions, decoder architectures, fusion strategies, attention modules, hybrid decoding designs, test-time augmentation, and exponential moving average training.

\noindent The main contributions of this work are:

\begin{itemize}[leftmargin=*, topsep=2pt, itemsep=1pt]
  \item A \textbf{dual-encoder multimodal architecture} that processes RGB and geophysical inputs using Swin Transformer V2 encoders and performs multi-scale feature fusion through lightweight concatenation-projection modules.
  
  \item A \textbf{decoder-agnostic} multimodal segmentation framework with test-time augmentation, where the main proposed model uses a \textbf{UNet++ decoder} and achieves the best mIoU, while a hybrid SegFormer--UNet++ decoder is evaluated separately in ablation.

  \item A comprehensive \textbf{ multi-axis ablation study}, providing insights into effective architectural choices for multimodal segmentation under small labeled training data constraints.
\end{itemize}

Our model achieves \textbf{0.867 mIoU (F1 = 0.905)} on the development benchmark and \textbf{0.783 mIoU (F1 = 0.800)} on the held-out test phase of the PBVS 2026 MARS-LS Challenge.

\section{Related Work}
\label{sec:related}
\textbf{Semantic Segmentation Architectures}
 has evolved from early fully convolutional networks\cite{long2015fully}, which enabled  pixel-wise prediction, to encoder--decoder models that better preserve spatial details through skip connections. Among them, U-Net \cite{ronneberger2015u} became influential because of its simple design and effective multiscale feature aggregation. Later methods improved context modeling and boundary refinement. DeepLabv3+ \cite{chen2018encoder} combined atrous spatial pyramid pooling with an encoder--decoder framework for sharper segmentation, while UNet++ \cite{zhou2018unet++} redesigned skip connections with nested dense pathways to reduce the semantic gap between encoder and decoder features. More recently, transformer-based models have shown strong results in dense prediction. Architectures such as SegFormer \cite{xie2021segformer} and Swin \cite{liu2022swin} show that hierarchical transformers can capture long-range context while preserving multiscale structure. Together, these advances motivate the use of transformer encoders with multiscale decoders for challenging segmentation tasks.

\textbf{Landslide Segmentation in Remote Sensing}
 remains difficult because of large intra-class variability, complex terrain, and limited annotated data. The Landslide4Sense benchmark \cite{ghorbanzadeh2022landslide4sense} introduced a standardized multimodal dataset and showed that combining optical imagery with terrain features such as DEM and slope improves performance. Building on this, recent studies have explored stronger multimodal fusion and better representation learning. For example, AST-UNet \cite{liu2024attention} enhances SwinUNet with attention mechanisms to improve boundary continuity and reduce noise. D2FLS-Net \cite{zhao2025d2fls} further proposes a dual-stage DEM-guided fusion transformer for more effective integration of terrain and image features. Another direction uses vision foundation models; TransLandSeg \cite{hou2024translandseg} adapts a pretrained transformer.\newline
\textbf{Martian Landslide Segmentation.}
Compared with terrestrial landslide detection, Martian landslide segmentation from orbital imagery is still relatively underexplored. One of the earliest works, MarsLS-Net \cite{paheding2024marsls}, introduced both a multimodal architecture and a benchmark dataset for this task. More recently, MMLSv2 \cite{paheding2026mmlsv2} expanded this line of work with a seven-band multimodal dataset. These studies show that Martian landslide detection differs from Earth-based settings in two major ways: labeled data are extremely limited, and useful information is spread across heterogeneous modalities such as high-resolution imagery, elevation, and thermal measurements. This makes Mars a strong testbed for architectures that can balance modality-specific representation learning with efficient multimodal fusion.

\section{Proposed Method}
\label{sec:method}

\begin{figure*}[t]
  \centering
  \includegraphics[width=\textwidth]{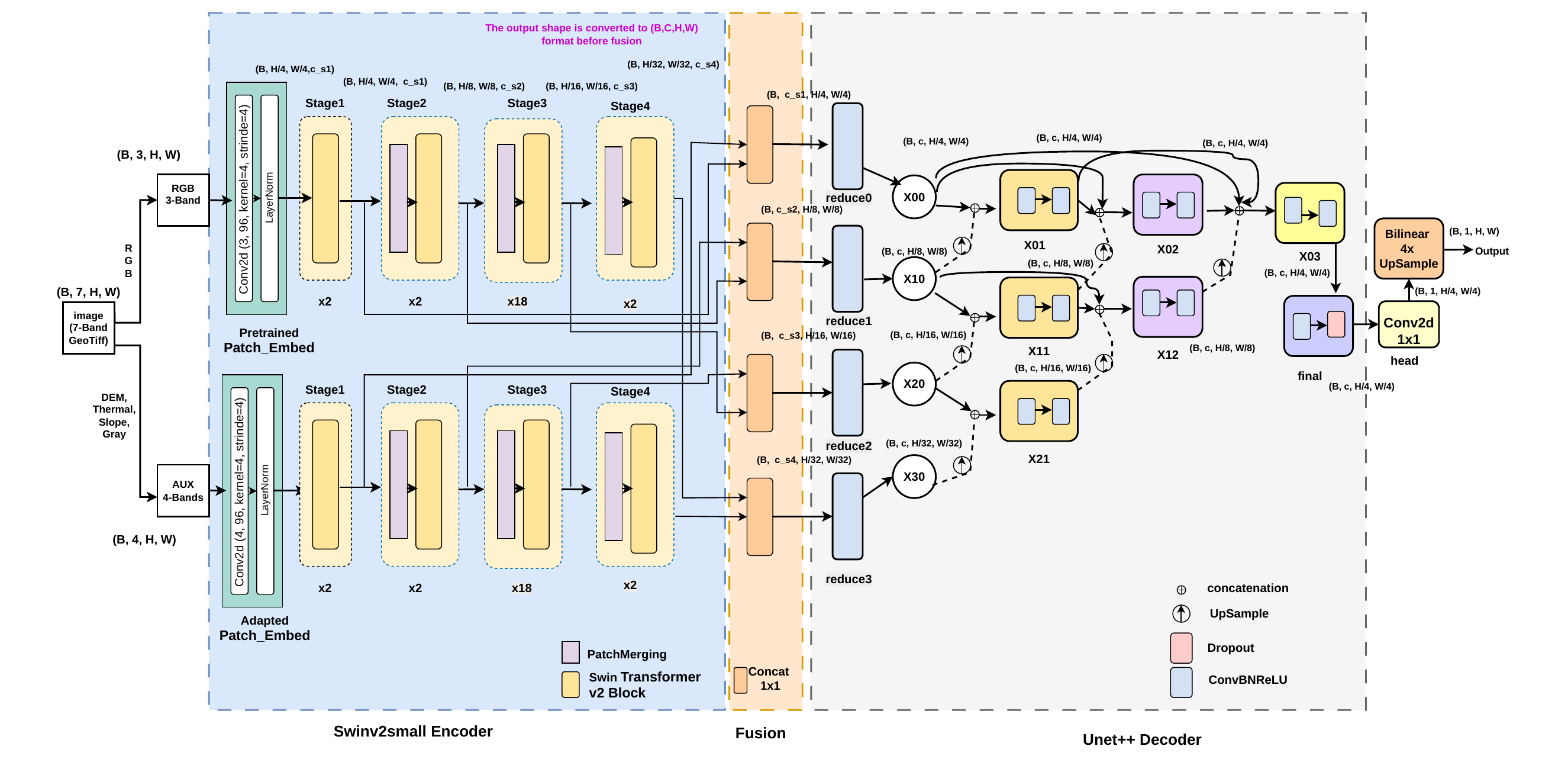}
\caption{DualSwinFusionSeg: two SwinV2-Small encoders process RGB and AUX inputs into four-level feature hierarchies; per-scale features are concatenated,reduced via $1\times1$ convolution, and decoded by UNet+,followed by a Conv\,$1\times1$ head and bilinear upsampling to the final segmentation mask.}
  \label{fig:arch}
\end{figure*}

We introduce DualSwinFusionSeg, a multimodal binary segmentation model for heterogeneous, multi-resolution Mars orbital data. As shown in Figure~\ref{fig:arch}, it comprises: (i) two hierarchical vision transformer encoders extracting scale-wise features from RGB and auxiliary geophysical inputs; (ii) a cross-modal fusion module integrating the features at each scale; and (iii) a dense decoder predicting the segmentation mask. Prediction after feature fusion can be written as:
\begin{equation}
\hat{y}=\sigma\!\left(\mathrm{Up}\!\left(\mathrm{Head}\!\left(\mathrm{UNet{+}{+}}\!\left(\{F^{l}\}_{l=1}^{4}\right)\right)\right)\right)
\end{equation}

where $F^l$ is the fused feature at scale $l$ and $\hat{y}\in[0,1]^{H\times W}$ is the per-pixel landslide probability.

\subsection{Input Representation}
\label{sec:input}
Each $128\times128$ GeoTIFF tile contains seven co-registered channels derived from five instruments across four Mars orbital missions. Rather than treating them as a single tensor, we split them into two semantically coherent groups. The RGB tensor $X^\text{rgb}\in\mathbb{R}^{B \times 3 \times H \times W}$ comprises bands 5–7 (red, green, blue) from the Viking colorized global mosaic (${\sim}232$\,m/pixel). The auxiliary tensor $X^\text{aux}\in\mathbb{R}^{B \times 4 \times H \times W}$ contains THEMIS thermal inertia (band~1), slope (band~2), DEM (band~3), and CTX
greyscale imagery (band~4), jointly encoding thermophysical properties, terrain geometry, and medium-resolution optical context (${\sim}6$–$200$\,m).
Both tensors are resized to a common input size $H\times W$ (default $128\times128$) and normalized in two steps: per-image, per-band percentile scaling ($P_1$--$P_{99}$ to $[0,1]$), followed by channel-wise standardization using dataset-level mean and standard deviation (Section~\ref{sec:impl}).

\subsection{Dual Swin Encoder}
\label{sec:encoders}

A single encoder struggles to reconcile RGB imagery and DEM/slope fields because their dynamic ranges, spatial frequencies, and pixel semantics differ. We therefore use two \emph{independent} Swin Transformer V2-Small (SwinV2-S) backbones~\cite{liu2022swin}, producing separate hierarchical feature representations for each modality before fusion.

\paragraph{Swin Transformer V2 backbone.}
SwinV2 computes self-attention in non-overlapping $8\times8$ windows, alternating regular and shifted windows for cross-window interaction. It uses log-spaced continuous position bias (Log-CPB)~\cite{liu2022swin} for relative positional encoding which generalizes better to unseen resolutions than the discrete bias of the original Swin~\cite{liu2021swin}. For $128\times128$ input, it outputs four stages with strides $\{4,8,16,32\}$, channels $\{96,192,384,768\}$, and map sizes $\{32\times32,16\times16,8\times8,4\times4\}$. Both encoders use \texttt{features\_only=True} and \texttt{out\_indices=(0,1,2,3)}. Although the checkpoint name is \texttt{swinv2\_small\_window8\_256}, we set \texttt{patch\_embed.img\_size=None} and \texttt{patch\_embed.strict\_img\_size=False}, enabling $128\times128$ inputs.

\vspace{-5mm}

\paragraph{Encoder initialization.}
Both encoders load ImageNet-21k pretrained weights of
\texttt{swinv2\_small\_window8\_256} from \texttt{timm}~\cite{wightman2023rwightman}, each initialized independently from the same checkpoint. For the AUX encoder, the patch-embedding convolution $W_0\in\mathbb{R}^{96\times3\times4\times4}$ is replaced by $W'\in\mathbb{R}^{96\times4\times4\times4}$: channels 1--3 copy $W_0$ directly, and channel~4 is set to the per-position RGB mean, $W'_{:,4,:,:}=\frac{1}{3}\sum_{c=1}^{3}W_{0,:,c,:,:}$. This \emph{channel-mean warm initialization} preserves pretrained spatial
filters for the first three channels while providing a reasonable starting point for the CTX greyscale channel (band~4); all other transformer block weights remain unchanged from the pretrained checkpoint.

\subsection{Cross-Modal Feature Fusion}
\label{sec:fusion}

Let $F_\text{RGB}^l, F_\text{aux}^l \in \mathbb{R}^{B \times C_l \times H_l \times W_l}$ denote the feature maps from the two encoders at pyramid scale $l\!\in\!\{1,2,3,4\}$, with $C_l \in \{96,192,384,768\}$. The streams are fused via channel-wise concatenation followed by $1\!\times\!1$ convolution:
\begin{equation}
F^l = f_\theta^l\big([F_\text{RGB}^l \;\| \; F_\text{aux}^l]\big),
\quad l = 1,\dots,4,
\label{eq:fusion}
\end{equation}
where $f_\theta^l$ is a $1\times1$ convolution that reduces the $2C_l$-channel tensor back to $C_l$. Four independent per-scale Conv($1\times1$)-BN-ReLU projections inside the decoder then map each scale to $C{=}256$ channels at the decoder input.
This design enables per-location channel mixing without spatial token interaction. Compared to cross-attention, it avoids the quadratic attention term $O(N^2C_l)$ and uses pointwise-convolution compute $O(NC_l^2)$ with $O(C_l^2)$ parameters per scale.

\subsection{Decoder Network and Classification Head}
\label{sec:decoder}

The fused feature pyramid $\{F^l\}_{l=1}^4$ is decoded by UNet++~\cite{zhou2018unet++}, which extends U-Net with \emph{dense nested skip connections} that progressively transform encoder features before fusion. Formally, at node $(i,j)$:
\begin{equation}
\mathbf{x}^{i,j}=\mathcal{H}_{i,j}\!\left(\mathrm{Cat}\!\left(\{\mathbf{x}^{i,k}\}_{k=0}^{j-1},\,\mathrm{Up}(\mathbf{x}^{i+1,j-1})\right)\right),\quad j\ge1,
\label{eq:unetpp}
\end{equation}

where $\mathbf{x}^{i,0}=\mathrm{Proj}_i(F^{i+1})$ denotes the projected encoder feature at scale $i\in\{0,1,2,3\}$, $\mathrm{Up}(\cdot)$ is bilinear upsampling, and $\mathcal{H}_{i,j}$ is a  block with two Conv(3$\!\times\!$3)-BatchNorm-ReLU layers. 
Deep supervision is disabled in default setting; only the final decoder output node (denoted $\mathbf{x}^{0,3}$) contributes to the loss. Prior to the dense nodes, four independent Conv($1\times1$)-BN-ReLU projections map each encoder scale $\{96,192,384,768\}$ to a common $C{=}256$-channel space. The final node $\mathbf{x}^{0,3}$ is refined by a Conv(3$\times$3)-BN-ReLU followed by Dropout ($p=0.1$), then a Conv(1$\times$1) head outputs a single logit map, bilinearly upsampled to expected output resolution($128\!\times\!128$). 

The network has \textbf{$\sim$113\,M parameters}: two SwinV2-S encoders (49\,M each), fusion layers (1.6\,M), and the UNet++ decoder (13.9\,M).

\subsection{Training Objective}
\label{sec:loss_fn}

We train the network with a composite loss combining class-weighted Binary Cross-Entropy (BCE) and soft Dice:
\begin{equation}
\mathcal{L} = \tfrac{1}{2}\mathcal{L}_\mathrm{BCE}(w^+) + \tfrac{1}{2}\mathcal{L}_\mathrm{Dice},
\label{eq:loss}
\end{equation}
where $w^+ = N_\text{neg}/N_\text{pos} \approx 1.86$ compensates for the $\sim$35\% foreground ratio, and $\hat{p}_i = \sigma(z_i)$ is the predicted probability. The soft Dice loss is
\begin{equation}
\mathcal{L}_\mathrm{Dice} = 1 - \frac{2\sum_i \hat{p}_i y_i }{\sum_i \hat{p}_i + \sum_i y_i + \varepsilon}, \quad \varepsilon=10^{-7}.
\label{eq:dice}
\end{equation}

BCE provides stable per-pixel gradients to prevent collapse to background, while Dice directly maximizes intersection-over-union, aligning training with mIoU evaluation. Ablations over ten alternative losses (Section~\ref{sec:loss_ablation}) show this combination consistently outperforms each term alone and other variants including focal, Tversky, soft-IoU, and boundary Dice losses. Table \ref{tab:training} shows choices of hyper parameters during training.

\section{Experimental Results}
\label{sec:experiments}

\subsection{PBVS 2026 MARS-LS Dataset}
The MARS-LS challenge dataset~\cite{marsls2026} is based on MMLSv2 and provides co-registered $128\times128$ GeoTIFF tiles with pixel-wise binary landslide masks. Each tile contains seven aligned channels: thermal inertia (band 1; THEMIS, $\sim$100\,m/pixel), slope (band 2; derived from DEM), DEM (band 3; MOLA blended with HRSC, $\sim$200\,m/pixel), grayscale imagery (band 4; CTX, $\sim$6\,m/pixel), and RGB basemap channels (bands 5-7; Viking colorized global mosaic, $\sim$232\,m/pixel). This heterogeneous, multi-source input motivates the dual-encoder design in Section~\ref{sec:method}.

\noindent \textbf{Dataset splits and class statistics.}
The official development data comprises 465 labelled training tiles and 66 labelled validation tiles. The positive-class weight, $w^+ = N_\mathrm{neg}/N_\mathrm{pos} \approx 1.86$, computed from the training tiles, accounts for moderate class imbalance, with foreground pixels averaging $\sim$35.4\% across the training set.

\begin{table}[t]
\centering
\caption{Training hyperparameters used across all experiments.}
\label{tab:training}
\small
\setlength{\tabcolsep}{4pt}
\begin{tabular}{@{}lc lc@{}}
\toprule
Hyperparameter & Value & Hyperparameter & Value \\
\midrule
Optimiser       & AdamW              & Epochs / fold  & 50 \\
Learning rate   & $2{\times}10^{-4}$ & Batch size     & 16 \\
Weight decay    & $10^{-4}$          & Precision      & FP16 \\
LR warm-up      & 3 epochs           & LR schedule    & Cosine \\
EMA decay ($\gamma$) & 0.995         & CV folds       & 5 \\
\bottomrule
\end{tabular}
\end{table}

\begin{figure}[t]
\centering
\includegraphics[width=\columnwidth]{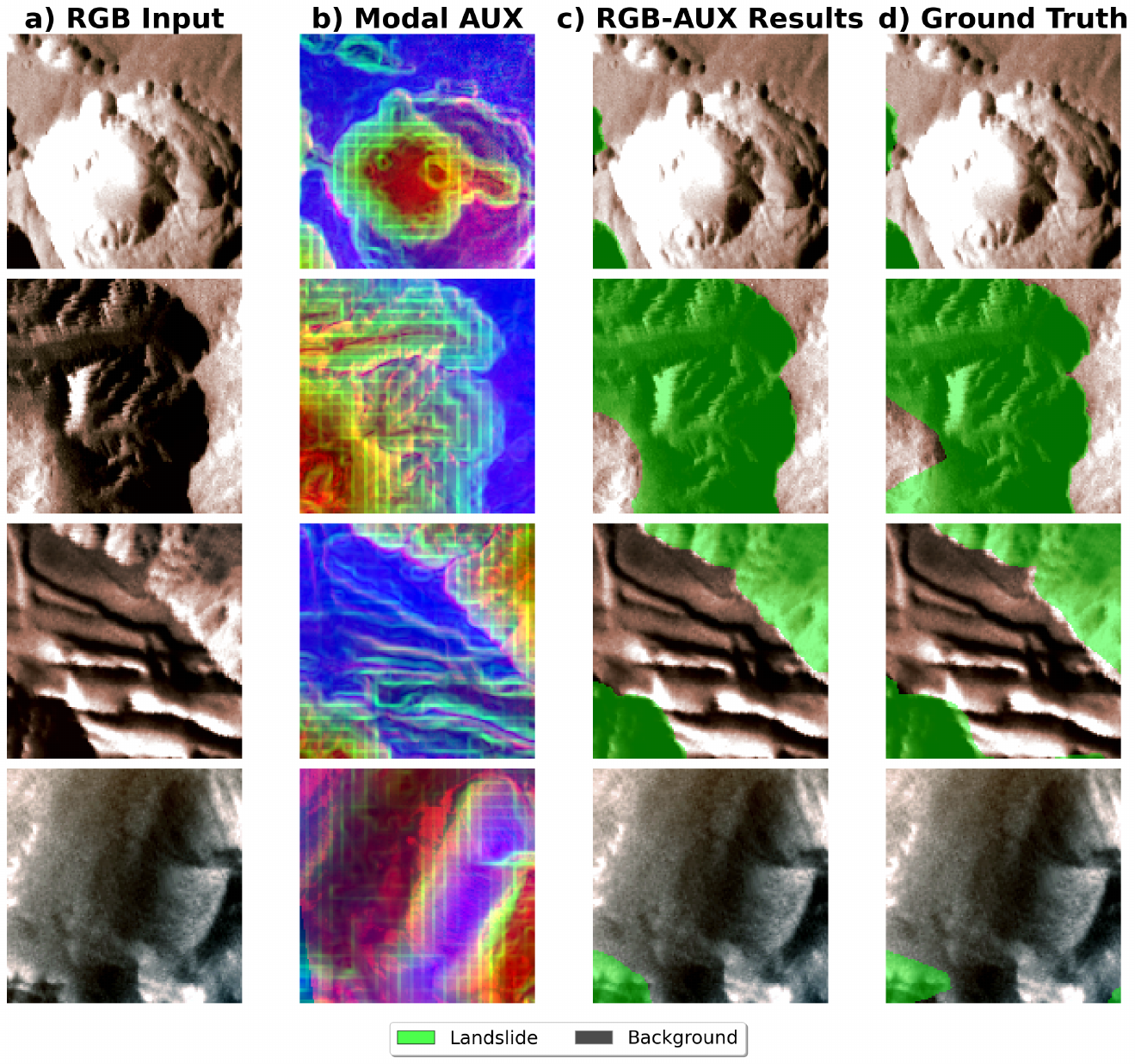}
\caption{Qualitative segmentation results. 
}
\label{fig:qualitative}
\end{figure}

\begin{table}[t]
\centering
\caption{Official CodaBench evaluation for the final submission.}
\label{tab:codabench}
\small
\setlength{\tabcolsep}{4pt}
\begin{tabular}{@{}lcccccc@{}}
\toprule
Phase & mIoU & IoU$_\text{fg}$ & IoU$_\text{bg}$ & F1 & Prec. & Recall \\
\midrule
Development & 0.867 & 0.827 & 0.907 & 0.905 & 0.900 & 0.911 \\
Test        & 0.783 & 0.677 & 0.890 & 0.800 & 0.760 & 0.860 \\
\bottomrule
\end{tabular}
\end{table}

\noindent \textbf{Preprocessing.}
A two-stage normalization pipeline is applied consistently during training and inference to ensure stable input statistics. \textit{Stage 1}: each tile-channel is clipped to its $[P_1, P_{99}]$ percentile range and linearly rescaled to $[0,1]$, mitigating sensor artifacts and extreme terrain outliers. \textit{Stage 2}: dataset-level z-score standardization, $x' = (x - \mu_c)/(\sigma_c + \epsilon)$, is applied using channel-wise statistics computed solely from the  training tiles.

\begin{table*}[t]
\centering
\caption{Progressive improvement across experimental configurations.
Metrics marked (\dag) are single-fold results on the competition validation
split; all other entries are official CodaBench leaderboard scores.
Configurations marked ($\star$) are novel architecture explorations that
provide design insights but do not surpass the final UNet++ configuration
on mIoU.}
\label{tab:progression}
\small
\setlength{\tabcolsep}{9pt}
\begin{tabular}{@{}lcc@{}}
\toprule
Configuration & mIoU & F1 \\
\midrule
SwinTiny + FPN, RGB + DEM (4 channels)\dag                                   & 0.672 & - \\
SwinTiny + FPN, all 7 channels\dag                                           & 0.743 & 0.821 \\
SwinTiny + FPN, class-weighted BCE + Dice\dag                                & 0.762 & 0.840 \\
MMSFormer~\cite{reza2024mmsformer} (B2+B1 encoders), all 7 channels\dag             & 0.716 & - \\
DualSwinTiny + FPN, gated cross-modal fusion\dag                             & 0.793 & 0.848 \\
SwinTiny + FPN (CodaBench development leaderboard)                           & 0.837 & 0.884 \\
DualSwinV2-S + UNet++, weighted-sum cross-modal fusion\dag                   & 0.824 & 0.868 \\
DualSwinV2-S + FPN, concatenation-based cross-modal fusion\dag               & 0.836 & 0.878 \\
$\star$ DualSwinV2-S + Hybrid Decoder (single-fold)\dag                      & 0.828 & \textbf{0.877} \\
$\star$ DualSwinV2-S + Hybrid Decoder + channel attention (5-fold CV)\dag    & $0.805\pm0.004$ & $0.872\pm0.004$ \\
\textbf{DualSwinV2-S + UNet++, 5-fold ensemble (development)}                & \textbf{0.867} & 0.905 \\
\textbf{DualSwinV2-S + UNet++, 5-fold ensemble (test)}                       & \textbf{0.783} & 0.800 \\
\bottomrule
\end{tabular}
\end{table*}

\subsection{Implementation Details}
\label{sec:impl}

\textbf{Hardware and software.}
All experiments were conducted on Kaggle computational notebooks using
4 vCPUs (Intel Xeon @ 2.00\,GHz), 31\,GiB system RAM, and
2$\times$ NVIDIA Tesla T4 GPUs (16\,GiB VRAM each; CUDA 13.0).
The implementation is based on Python 3.12, PyTorch 2.9+cu126, the
\texttt{timm} model library~\cite{wightman2023rwightman} for SwinV2 backbone loading,
Albumentations~\cite{buslaev2020albumentations} for online data augmentation,
and \texttt{rasterio} for GeoTIFF input/output.

\noindent \textbf{Data augmentation.}  
To mitigate overfitting on the limited training set, online augmentation is applied during training. Geometric transformations are applied jointly to all input channels and the segmentation mask to preserve spatial consistency. Photometric transformations are restricted to RGB, as altering geophysical channels (DEM, Slope, Thermal) would distort their physically meaningful values (Table~\ref{tab:augmentation}).
\begin{table}[t]
\centering
\caption{Online data augmentation pipeline applied during training.}
\label{tab:augmentation}
\small
\setlength{\tabcolsep}{3pt}
\begin{tabular}{@{}lp{2.2cm}p{3.0cm}@{}}
\toprule
Type & Transform & Setting \\
\midrule
\multirow{3}{*}{Geometric}
  & Horizontal / Vertical flip   & $p = 0.5$ each \\
  & Random rotation($90^\circ$) & $p = 0.5$ \\
  & Affine transform             & $\pm5\%$ shift, $[0.9,1.1]\!\times$ scale, $\pm20^\circ$, $p{=}0.5$ \\
\midrule
\multirow{3}{*}{\parbox{1.4cm}{Photometric\\(RGB only)}}
  & Gaussian blur            & kernel $3$--$7$, $\sigma$ auto, $p{=}0.15$ \\
  & Brightness \& Contrast  & limit $\pm0.2$ each, $p{=}0.3$ \\
\bottomrule
\end{tabular}
\end{table}

\noindent \textbf{Optimisation.}  
All models are trained with the AdamW optimizer~\cite{loshchilov2017decoupled} under mixed-precision (FP16). A 3-epoch linear learning-rate warm-up is followed by cosine annealing for the remaining epochs. An exponential moving average (EMA) of model weights with decay $\gamma = 0.995$ is maintained throughout training. After each epoch, both instantaneous and EMA weights are evaluated on the held-out fold, and the checkpoint with higher mIoU is retained.

\noindent \textbf{Ensemble inference and test-time augmentation.}  
For inference, the five fold-best checkpoints predictions are combined via ensemble averaging: each test tile is processed by all five models, and the resulting probability maps are averaged pixel-wise. Four-view test-time augmentation (TTA) is applied, processing the original tile and three transformed versions (horizontal flip, vertical flip, $90^\circ$ rotation) through the ensemble, producing $5 \!\times\! 4 = 20$ probability maps per tile. These maps are averaged and thresholded at $\tau = 0.51$ to yield the final binary mask.

\subsection{Main Results}
\label{sec:final}

Table~\ref{tab:codabench} presents the official metrics from the PBVS 2026
CodaBench evaluation server for the submitted model. 
The final model achieves mIoU\,=\,0.867 and F1\,=\,0.905 on the
development phase, and mIoU\,=\,0.783 and F1\,=\,0.800 on the
held-out test phase.  The development score notably exceeds the 5-fold
cross-validation mean (mIoU\,=\,0.84) primarily because the 20-prediction
ensemble (5~folds $\times$ 4~TTA views) combined with EMA weight averaging
substantially reduces prediction variance relative to any single fold
checkpoint.

The $\Delta\mathrm{mIoU}\,{=}\,0.084$ gap between development and test
phases is consistent with domain shift rather than training-set overfitting.
Notably, the precision drop
($0.900\!\to\!0.760$) exceeds the recall drop ($0.911\!\to\!0.860$),
indicating that the model retains sensitivity to landslide deposits but
generates more false positives on geologically similar features in the test domain.
Table~\ref{tab:progression} traces the step-by-step mIoU improvement
from the initial baseline to the final submission, demonstrating a
consistent progression driven by each architectural and training decision.

Figure~\ref{fig:qualitative} shows representative qualitative results, highlighting  predicted masks against ground truth.

\subsection{Ablation Studies}
\label{sec:ablation}

A systematic ablation study was conducted to isolate the contribution of
each design decision. Each ablation varies a single component while
holding all others fixed at their best-known settings.

\subsubsection{Ablation study on modality combinations.}
\label{sec:modality_ablation}
A SwinTiny+FPN model ($\sim$33\,M parameters) is trained with different
subsets of the seven input channels to quantify the marginal contribution
of each sensing modality (Table~\ref{tab:modality}).
\begin{table}[H]
\centering
\caption{Ablation study on modality combinations.}
\label{tab:modality}
\small
\setlength{\tabcolsep}{4pt}
\begin{tabular}{@{}p{3.4cm}ccc@{}}
\toprule
Modalities & Channels & Val mIoU & Val Dice \\
\midrule
RGB + DEM           & 4 & 0.672 & 0.758 \\
RGB + Thermal       & 4 & 0.661 & 0.750 \\
RGB + DEM + Slope   & 5 & 0.721 & 0.798 \\
RGB + DEM + Thermal & 5 & 0.691 & 0.778 \\
All 7 channels      & 7 & \textbf{0.743} & \textbf{0.821} \\
\bottomrule
\end{tabular}
\end{table}
Terrain geometry channels (DEM and Slope) contribute most to segmentation
accuracy.  A typical Martian landslide exhibits a steep source scarp (high
slope magnitude) directly adjacent to a gently inclined depositional fan
(low slope magnitude)--a morphological signature that is largely
albedo-independent and therefore detectable even when the RGB channels
show low contrast.  Thermal inertia provides a weaker discriminative
signal at its native $\sim$100\,m resolution, but its combination with
the remaining channels reduces false positives on rocky outcrops, which
share high slope values with landslide scarps but differ in thermal
properties.  The full 7-channel configuration achieves the highest mIoU
and Dice, confirming that each sensing modality contributes independent
and complementary information.

\subsubsection{Ablation study on loss fucntion}
\label{sec:loss_ablation}

\begin{table}[t]
\centering
\caption{Ablation study on the loss function.}
\label{tab:loss}
\small
\setlength{\tabcolsep}{4pt}
\begin{tabular}{@{}p{0.55cm}p{3.2cm}cc@{}}
\toprule
ID & Loss function & Val mIoU & Val Dice \\
\midrule
$\mathcal{L}_1$ & BCE                                           & 0.728 & 0.806 \\
$\mathcal{L}_2$ & Dice                                          & 0.752 & 0.833 \\
$\mathcal{L}_3$ & BCE + Dice                                    & 0.758 & 0.833 \\
$\mathcal{L}_4$ & BCE($w^+$) + Dice                             & \textbf{0.762} & \textbf{0.840} \\
$\mathcal{L}_5$ & Focal~\cite{lin2017focal}                & 0.744 & 0.818 \\
$\mathcal{L}_6$ & Focal + Dice                                  & 0.748 & 0.827 \\
$\mathcal{L}_7$ & Soft-IoU                                      & 0.779 & 0.851 \\
$\mathcal{L}_8$ & BCE + Soft-IoU                                & 0.751 & 0.827 \\
$\mathcal{L}_9$ & Tversky ($\alpha\!=\!0.3$)~\cite{salehi2017tversky} & 0.747 & 0.827 \\
$\mathcal{L}_{10}$ & Dice + Boundary Dice                       & 0.704 & 0.790 \\
$\mathcal{L}_{11}$ & Dice + Lovasz ~\cite{berman2018lovasz}      & 0.757 & 0.829 \\
$\mathcal{L}_{12}$ & BCE($w^+$)+Dice+Lovasz                          & 0.760 & 0.846 \\

\bottomrule
\end{tabular}
\end{table}

$\mathcal{L}_4$ (class-weighted BCE combined with soft Dice) achieves the
best mIoU and the lowest fold-to-fold variance (Table \ref{tab:loss}).  While Soft-IoU
($\mathcal{L}_7$) records a higher single-fold peak, it shows substantially
greater instability across folds ,a well-known characteristic of loss
functions whose gradients saturate near the optimum when training sets are
small~\cite{buslaev2020albumentations}.  Boundary Dice
($\mathcal{L}_{10}$) performs worst because up-weighting boundary pixels
amplifies annotation noise present at the deposit perimeters in this
dataset.  The complementary roles of the two terms BCE providing
per-pixel gradient stability and Dice providing a direct mIoU
proxy justify the equal-weight combination selected for all subsequent
experiments.

\subsubsection{Baseline and dual-encoder}
\label{sec:baselines}

The MMSFormer baseline~\cite{reza2024mmsformer} achieves a validation mIoU of
0.716--4.6 percentage points below the SwinTiny+FPN single-encoder
baseline (0.762).  MMSFormer's cross-modal transformer attention was
designed and evaluated on indoor RGB-D scenes where modality co-occurrence
is highly structured; its inductive biases do not transfer to planetary
multi-spectral data where the seven channels arise from physically
unrelated sensing phenomena.  Separating feature extraction into two
independent encoder streams (DualSwinTiny + FPN with gated cross-modal
fusion) raises validation mIoU to 0.793, a gain of $+$3.1 percentage
points over the single-encoder SwinTiny+FPN baseline.  This confirms that
modality-specific feature extraction, performed prior to any cross-modal
integration, is the primary driver of Stage II improvement.

\begin{table}[H]
\centering
\caption{Ablation study on the decoder choice.}
\label{tab:decoder}
\small
\setlength{\tabcolsep}{5pt}
\begin{tabular}{@{}lcc@{}}
\toprule
Decoder & Val mIoU & Val F1 \\
\midrule
FPN~\cite{lin2017feature}                      & 0.800 & 0.849 \\
UPerNet~\cite{xiao2018unified}             & 0.818 & 0.865 \\
SegFormer MLP~\cite{xie2021segformer}      & 0.806 & 0.857 \\
DeepLabV3+~\cite{chen2018encoder}       & 0.801 & 0.846 \\
UNet++~\cite{zhou2018unet++}               & \textbf{0.824} & \textbf{0.868} \\
\bottomrule
\end{tabular}
\end{table}

\begin{table}[t]
\centering
\caption{Ablation study on cross-modal fusion strategy.}
\label{tab:fusion}
\small
\setlength{\tabcolsep}{4pt}
\begin{tabular}{@{}llcc@{}}
\toprule
Decoder & Cross-modal fusion & mIoU & F1 \\
\midrule
FPN     & Concatenation-based & \textbf{0.836} & \textbf{0.878} \\
UPerNet & Concatenation-based & 0.822 & 0.867 \\
UNet++  & Concatenation-based & 0.814 & 0.857 \\
UNet++  & Weighted-sum        & 0.803 & 0.850 \\
UPerNet & Weighted-sum        & 0.824 & 0.868 \\
\bottomrule
\end{tabular}
\end{table}

\subsubsection{Decoder and cross-modal fusion}
\label{sec:decoder_ablation}

Two complementary ablations are conducted using the larger SwinV2-Small
backbone ($\sim$113 \,M total parameters): (i)~a decoder ablation with
fixed weighted-sum cross-modal fusion comparing five decoder architectures
(Table~\ref{tab:decoder}); and (ii)~a fusion ablation comparing two
cross-modal fusion strategies across multiple decoder choices
(Table~\ref{tab:fusion}).

UNet++ achieves the highest mIoU in the decoder-only comparison owing to
its dense nested skip pathways, which allow boundary-level cues at every
encoder scale to propagate directly to the output without being bottlenecked
by a single skip connection per resolution level.  Although FPN records
a marginally higher mIoU with concatenation-based fusion in
Table~\ref{tab:fusion}, UNet++ exhibits lower fold-to-fold variance in
5-fold cross-validation, making it the more reliable choice for the final
submission.  Concatenation-based cross-modal fusion consistently surpasses
weighted-sum fusion across all tested decoders ($+$0.011 mIoU on average),
confirming that a learned full-rank linear interaction between RGB and
auxiliary feature spaces is more expressive than a fixed scalar weighting.

\subsubsection{Architecture explorations}
\label{sec:novel_exploration}

\noindent \textbf{Hybrid SegFormer-UNet++ Decoder.}
Analysis of per-tile predictions from the Stage III UNet++ model reveals
a recurring failure mode: large, spatially diffuse landslide deposits are
segmented incompletely, with isolated foreground fragments rather than
continuous connected regions.  This is attributed to the local nature of
UNet++ skip connections, which prioritise high-frequency boundary cues
over long-range scene context.  A Hybrid Decoder is therefore constructed
by running a UNet++ graph and a SegFormer MLP head~\cite{xie2021segformer}
in parallel on the shared fused feature pyramid.  
The UNet++ path captures fine-grained boundary detail via dense skip connections, while the SegFormer path projects all four pyramid levels to a shared channel width with independent $1\times1$ convolutions, upsamples them to the finest scale, and aggregates them through a single $1\times1$ fusion, improving recall on large, diffuse deposits. The two decoder outputs are aligned to the same spatial
resolution and 
fused by a learned $1\times1$ conv + BN-ReLU + SE gate before the classification head.

\begin{table}[H]
\centering
\caption{Hybrid decoder vs.\ plain UNet++.}
\label{tab:hybrid}
\small
\setlength{\tabcolsep}{3pt}
\begin{tabular}{@{}lcccccc@{}}
\toprule
Decoder & mIoU & IoU$_\text{fg}$ & IoU$_\text{bg}$ & Prec. & Recall & F1 \\
\midrule
UNet++ & \textbf{0.836} & \textbf{0.776} & \textbf{0.895} & \textbf{0.895} & 0.850 & 0.872 \\
Hybrid & 0.828          & 0.769          & 0.887          & 0.856          & \textbf{0.899} & \textbf{0.877} \\
\bottomrule
\end{tabular}
\end{table}
Table~\ref{tab:hybrid} quantifies the precision-recall trade-off in a
controlled single-fold evaluation.  The Hybrid Decoder achieves a higher
F1 score (0.877 vs.\ 0.872) by raising recall from 0.850 to 0.899
through the SegFormer global path, at the cost of a moderate precision
reduction (0.895 to 0.856) and a small mIoU penalty (0.836 to 0.828).
Since mIoU is the primary MARS-LS ranking metric, plain UNet++ is used for final submission. However, when F1 or coverage recall is prioritized, the Hybrid Decoder is preferred.

\begin{table*}[t]
\centering
\caption{Ablation study on channel attention.
IA\,=\,input-level attention; IE\,=\,intra-encoder attention;
DO\,=\,decoder-output attention.}
\label{tab:attention}
\small
\setlength{\tabcolsep}{7pt}
\begin{tabular}{@{}lccclcc@{}}
\toprule
Config. & IA & IE & DO & Attention-gated fusion & CV mIoU & CV F1 \\
\midrule
E10-a & -  & ECA & -   & ECA-gated  & $0.806 \pm 0.009$ & $0.873 \pm 0.009$ \\
E10-b & -  & ECA & -   & CBAM-gated & $0.805 \pm 0.008$ & $0.872 \pm 0.009$ \\
E11   & ECA & ECA & CBAM & CBAM-gated & $0.802 \pm 0.011$ & $0.871 \pm 0.010$ \\
E12   & ECA & ECA & ECA  & ECA-gated  & $0.805 \pm 0.008$ & $0.872 \pm 0.008$ \\
E13   & ECA & ECA & SE   & SE-gated   & $0.803 \pm 0.006$ & $0.871 \pm 0.007$ \\
E9    & ECA & SE  & CBAM & ECA-gated  & $0.805 \pm 0.008$ & $0.872 \pm 0.008$ \\
\bottomrule
\end{tabular}
\end{table*}

\begin{table}[t]
\centering
\caption{Ablation study on test-time augmentation.}
\label{tab:tta}
\small
\setlength{\tabcolsep}{6pt}
\begin{tabular}{@{}lccc@{}}
\toprule
TTA Strategy & Val mIoU & $\Delta$mIoU & Inf.\ time \\
\midrule
No TTA                            & 0.8278 & $+$0.0000 & 0.9\,s \\
Horizontal flip                   & 0.8272 & $-$0.0006 & 1.0\,s \\
Vertical flip                     & 0.8276 & $-$0.0002 & 1.0\,s \\
\textbf{H-flip + V-flip (2-view)} & \textbf{0.8283} & $+$0.0005 & 1.2\,s \\
4-view ($90^\circ$ rotations)     & 0.8271 & $-$0.0007 & 1.2\,s \\
8-view (D4 symmetry group)        & 0.8231 & $-$0.0047 & 1.5\,s \\
\bottomrule
\end{tabular}
\end{table}

\begin{table}[t]
\centering
\caption{EMA vs.\ standard training.}
\label{tab:ema}
\small
\setlength{\tabcolsep}{8pt}
\begin{tabular}{@{}lccc@{}}
\toprule
Training configuration & Val mIoU & Val F1 & Best epoch \\
\midrule
EMA ($\gamma = 0.995$) & \textbf{0.8278} & \textbf{0.8692} & 28 \\
Standard (no EMA)      & 0.8178          & 0.8611          & 12 \\
\bottomrule
\end{tabular}
\end{table}

\noindent \textbf{Channel Attention Mechanisms.}
Building on the Hybrid Decoder, the effect of channel attention modules
is investigated by inserting them at three positions: at the input level
prior to the encoders (IA), within the encoder feature hierarchy (IE),
and at the decoder output (DO).  Three attention mechanisms are compared:
Efficient Channel Attention (ECA)~\cite{wang2020eca}, Squeeze-and-Excitation
(SE)~\cite{hu2018squeeze}, and Convolutional Block Attention Module
(CBAM)~\cite{woo2018cbam}.  Table~\ref{tab:attention} reports 5-fold
cross-validation mIoU and F1.

All attention configurations achieve CV mIoU in the narrow range
0.802-0.806; a spread of only 0.004 points, smaller than the
across-fold standard deviation of most runs.  This tight clustering
indicates that, given a training set of 465 tiles, the additional
parameters introduced by channel attention modules do not provide a
generalizable signal beyond the base concatenation-based fusion.
Input-level gating (E11-E13) performs marginally worse than
encoder-level gating alone (E10-a, E10-b), suggesting that applying
attention prior to the dual encoders partially disrupts the
modality-specific feature statistics that the dual-encoder design is
intended to preserve.  Channel attention is expected to provide
consistent gains when the training set is substantially larger
(${\gtrsim}10$k tiles) and when modality quality varies substantially
across samples.

\noindent \textbf{Late Fusion Ablation.}
We compare our proposed mid-fusion architecture against a late-fusion variant  that maintains separate encoder-decoder pipelines per modality and merges predictions via a learnable $\alpha$-blend: $\hat{y} = \sigma(\alpha)\cdot L_{\text{rgb}} + (1-\sigma(\alpha))\cdot L_{\text{aux}}$.
Despite doubling the decoder parameters, late fusion yields only a marginal mIoU difference ($0.8415 \pm 0.008$ vs.\ $0.8421 \pm 0.007$), well within the standard-deviation overlap across 5-fold CV.
Notably, the learned blending weight converges to $\alpha_{\text{rgb}} \approx 0.48$ across all folds, indicating near-equal modality reliance with a slight preference for auxiliary channels (DEM, slope), while the precision--recall trade-off shifts toward fewer false positives ($+1$ pp precision) at the cost of reduced recall ($-0.9$ pp).

\subsubsection{Inference - Test-Time Augmentation Ablation}
\label{sec:tta_ablation}

On a single model, TTA provides marginal gains or slight degradation (Table~\ref{tab:tta})
depending on the transform type.  Rotation-based augmentations reduce accuracy because they alter the orientation of
DEM gradient and slope features; the model has been trained on orbital tiles and learns direction-sensitive filters for the
typical slope of equatorial Martian terrain.  Horizontal and
vertical flips are approximately equivariant to the morphological landslide
signature and yield marginal improvement.  In the 5-fold ensemble setting,
4-view TTA is applied to reduce inter-fold prediction variance across the
20-model average, where the noise-averaging effect over independent model
predictions outweighs the marginal single-model degradation.

\subsubsection{Training - Exponential Moving Average Ablation}
\label{sec:ema_ablation}

EMA yields +0.010 mIoU and shifts best checkpoint from epoch 12 to 28
(Table~\ref{tab:ema}). Without EMA, validation
mIoU peaks early and then fluctuates as late-stage gradient updates produce
high-frequency oscillations around the loss minimum--a well-known
pathology of training on small datasets.  The EMA model is robust to these
short-term perturbations because it tracks a long-horizon parameter average:
a decay factor $\gamma = 0.995$ corresponds to an effective window of
$1/(1-\gamma) = 200$ update steps, approximately 6.7 epochs at the training
batch size of 16 on 465 tiles. This allows the network to continue
benefiting from informative gradient updates before the checkpoint is frozen.

\section{Discussion and Future Direction}
\label{sec:discussion}

Our dual-encoder architecture preserves modality-specific inductive biases by preventing the premature merging of fundamentally different features, isolating RGB visual features from the low-level gradient statistics of DEM and Slope channels. This separation yields a $+2.6\%$ mIoU improvement over single-encoder baselines and a $+4.1\%$ gain over MMSFormer \cite{reza2024mmsformer}. Additionally, on our limited dataset , static concat fusion avoids the overfitting inherent to ECA-gated fusion, which introduces ${\sim}1$M parameters per scale. We also identify a precision-recall tradeoff between decoder architectures. The dense nested skip connections of UNet++ propagate fine spatial details for precise boundary localization, maximizing mIoU for the MARS-LS competition. In contrast, our Hybrid SegFormer-UNet++ Decoder aggregates global context to improve the recall. This results in a superior F1 score with only a minor mIoU penalty ($-0.8\%$), making the Hybrid model the recommended architecture for recall-critical applications like hazard inventories. Despite strong development metrics (mIoU=0.867), our model exhibits a generalization gap on the held-out test phase (mIoU=0.783, $\Delta$=0.084). Drops in both precision ($0.900\!\to\!0.760$) and recall ($0.911\!\to\!0.860$) indicate domain shift caused by target tiles originating from a different distribution. While our 5-fold ensemble and EMA weight averaging mitigate variance, the model still struggles with structural ambiguities unresolvable at a $128\!\times\!128$-pixel context, including flat dust-covered albedos, zero-slope ancient landslides, and circular crater rims mimicking fresh scarps. Future work will address these spatial limitations  and cross-instrument domain adaptation to align features at their native resolutions. Scaling the Hybrid Decoder to larger datasets (${>}10$k tiles) will improve generalization without overfitting.

\section{Conclusion}
\label{sec:conclusion}

We presented \textbf{DualSwinFusionSeg}, a dual-encoder architecture for Martian landslide segmentation from seven-band orbital
remote sensing imagery. The model processes RGB and auxiliary geophysical modalities using separate encoders, fuses the resulting
multi-scale feature pyramids through lightweight concat modules, and reconstructs boundary-sensitive segmentation masks using a UNet++ decoder.
Through a 13-configuration ablation study, we show that separating RGB and geophysical streams from the earliest layers substantially improves
performance, providing a $+2.6\%$ mIoU gain over single-encoder fusion. We also find that simple concat–project fusion is more effective than
gated alternatives. While a Hybrid SegFormer-UNet++ decoder achieves the best F1 score by balancing
precision and recall, the plain UNet++ configuration yields the highest mIoU, highlighting a trade-off between segmentation metrics. The final model achieves
\textbf{0.867 mIoU} and \textbf{0.905 F1} on the PBVS 2026 MARS-LS development benchmark, and \textbf{0.783 mIoU} on the held-out test set,
demonstrating the effectiveness of modality-specific encoding and lightweight multimodal fusion for planetary surface segmentation.

\end{document}